\documentclass{article}

\usepackage{microtype}
\usepackage{graphicx}
\usepackage{subcaption}
\usepackage{multirow}
\usepackage{booktabs} % for professional tables
\usepackage{comment}
\usepackage{colortbl}

\usepackage{hyperref}
\newcommand{\Var}{\mathrm{Var}}

\usepackage{multirow}
\usepackage[table]{xcolor} % Required for row highlighting

\usepackage[preprint]{icml2026}

\usepackage{amsmath}
\usepackage{amssymb}
\usepackage{mathtools}
\usepackage{amsthm}

\usepackage[capitalize,noabbrev]{cleveref}

\theoremstyle{plain}

\theoremstyle{definition}

\theoremstyle{remark}

\icmltitlerunning{ Mixture-of-Gaussians with Uncertainty-based Gating}

\begin{document}

\twocolumn[
  \icmltitle{
MoGU: Mixture-of-Gaussians with Uncertainty-based Gating for Time Series Forecasting}



  \icmlsetsymbol{equal}{*}

  \begin{icmlauthorlist}
    \icmlauthor{Gilad Aviv}{yyy}
    \icmlauthor{Jacob Goldberger}{yyy}
    \icmlauthor{Yoli Shavit}{yyy}
  \end{icmlauthorlist}

 \icmlaffiliation{yyy}{Faculty of Engineering, Bar Ilan University, Ramat-Gan, Israel}
  \icmlcorrespondingauthor{Yoli Shavit}{yoli.shavit@biu.ac.il}

  \vskip 0.3in
]



\printAffiliationsAndNotice{}  % no special notice (required even if empty)

\begin{abstract}
   %We introduce Mixture-of-Gaussians with Uncertainty-based Gating (MoGU), a novel Mixture-of-Experts (MoE) framework designed for regression tasks and applied to time series forecasting. Unlike conventional MoEs that provide only point estimates, MoGU models each expert's output as a Gaussian distribution. This allows it to directly quantify both the forecast (the mean) and its inherent uncertainty (variance). MoGU's core innovation is its uncertainty-based gating mechanism, which replaces the traditional input-based gating network by using each expert's estimated variance to determine its contribution to the final prediction. Evaluated across diverse time series forecasting benchmarks, MoGU consistently outperforms single-expert models and traditional MoE setups. It also provides well-quantified, informative uncertainties that directly correlate with prediction errors, enhancing forecast reliability. Our code is available in the supplementary materials.
We introduce Mixture-of-Gaussians with Uncertainty-based Gating (MoGU), a novel Mixture-of-Experts (MoE) framework designed for regression tasks. MoGU replaces standard learned gating with an intrinsic routing paradigm where expert-specific uncertainty serves as the native gating signal. By modeling each prediction as a Gaussian distribution, the system utilizes predicted variance to dynamically weight expert contributions. We validate MoGU on multivariate time-series forecasting, a domain defined by high volatility and varying noise patterns. Empirical results across multiple benchmarks, horizon lengths, and backbones demonstrate that MoGU consistently improves forecasting accuracy compared to traditional MoE. Further evaluation via conformal prediction indicates that our approach yields more efficient prediction intervals than existing baselines. These findings highlight MoGU’s capacity for providing both competitive performance and reliable, high-fidelity uncertainty quantification. Our code is available at: \url{https://github.com/yolish/moe_unc_tsf}%Our code is available in the supplementary materials.
   
\end{abstract}

\section{Introduction}\label{sec:introduction}
Mixture-of-Experts (MoE) is an architectural paradigm that adaptively combines predictions from multiple neural modules, known as "experts," via a learned gating mechanism. This concept has evolved from ensemble-based MoEs, where experts, jointly trained with a gating function, are often full, independent models whose outputs are combined to improve overall performance and robustness \citep{jacobs1991adaptive}. More recently, MoE layers have been integrated within larger neural architectures, with experts operating in a latent domain. These "latent MoEs" offer significant scalability benefits, especially in large language models (LLMs) \citep{shazeer2017outrageously,fedus2022switch}.
MoE makes it possible to train massive but efficient LLMs, where each token activates only a fraction of the model’s parameters, enabling specialization, better performance, and lower computational cost compared to equally sized dense models.

Regardless of their specific implementation, conventional MoE systems typically produce point estimates, lacking a direct quantification of their uncertainty. In critical applications, this absence of uncertainty information hinders interpretability, making it difficult for users to gauge the reliability of a prediction and limits informed decision-making, as the system cannot express its confidence or identify ambiguous cases. Importantly, the learned gating mechanism, which dictates the relative contribution of each expert, does not take into account expert confidence, potentially leading to suboptimal routing decisions.

In this work, we propose Mixture-of-Gaussians with Uncertainty-based Gating (MoGU), a framework that reimagines the MoE architecture by centering expert coordination around predictive confidence. While traditional MoEs rely on auxiliary gating modules to learn routing weights, MoGU derives its gating logic directly from the experts' internal uncertainty. By modeling each expert's output as a Gaussian distribution, we utilize the predicted variance as a native proxy for confidence. This shift enables a self-aware routing mechanism where more certain experts naturally exert greater influence, effectively bypassing the need for separate, input-based gating functions.

We demonstrate the efficacy of MoGU on multivariate time-series forecasting, a domain characterized by non-stationarity and heteroscedastic noise. MoGU consistently yields superior predictive accuracy over standard learned-routing baselines across diverse time-series benchmarks and expert backbones. Furthermore, MoGU provides well-calibrated uncertainty estimates that exhibit a statistically significant positive correlation with empirical error ($p < 10^{-5}$). Through conformal prediction, we show that MoGU yields tighter, more efficient predictive intervals than traditional models, offering a robust solution for high-stakes time-series applications. %Beyond performance, we find that our confidence-based routing naturally prevents expert collapse by ensuring balanced expert utilization without auxiliary load-balancing terms. 
Finally, a comprehensive ablation study validates our key design choices, including the gating mechanism, head architecture, temporal resolution, and loss formulation.

In summary, our contributions are as follows:
\begin{itemize}
    \item \textbf{Uncertainty-Driven Gating Mechanism}: We introduce a novel routing paradigm that replaces conventional gating modules with a logic based on intrinsic expert confidence. This allows the model to dynamically prioritize experts based on their self-reported predictive uncertainty.
    \item \textbf{Improved Forecasting Accuracy:} We show that MoGU consistently reduces forecasting error across various benchmarks, horizon lengths, and expert architectures.
    \item \textbf{Well-Calibrated Uncertainty Estimation}:  By applying conformal prediction, we validate that MoGU provides superior predictive efficiency, producing narrower and more reliable intervals than standard baselines. %We further show that MoGU’s uncertainty signals are highly informative and positively correlated with actual error, at the expert and whole-model levels.
\end{itemize}

\section{Related Work}
\label{related_work}
\textbf{MoE Models.}
The pursuit of increasingly capable and adaptable artificial intelligence systems has led to the development of sophisticated architectural paradigms, among which the Mixture-of-Experts (MoE) stands out. MoE is an architectural concept that adaptively combines predictions from multiple specialized neural modules, often sharing a common architecture, through a learned gating mechanism. This paradigm allows for a dynamic allocation of computational resources, enabling models to specialize on different sub-problems or data modalities. Early implementations of MoE \citep{jacobs1991adaptive} focused on ensemble learning (ensemble MoE), where multiple models (experts) contributed to a final prediction. More recently, MoE layers have been seamlessly integrated within larger neural architectures, with experts operating in latent domains (latent MoE)  \citep{shazeer2017outrageously,fedus2022switch}. This integration has proven particularly impactful in the realm of large language models (LLMs), where MoE layers have been instrumental in scaling models to unprecedented sizes while managing computational costs \citep{lepikhin2020gshard, jiang2024mixtral, dai2024deepseekmoe}. By selectively activating only a subset of experts for each input token, MoEs enable models with vast numbers of parameters to achieve high performance without incurring the prohibitive inference costs of densely activated large models.
Despite  their contribution and adoption, both ensemble and latent MoE architectures typically output point estimates, both at the level of the individual expert and at the level of the overall model. This limits the ability to quantify uncertainty which is important for 
decision-making. 
Few works have explored uncertainty estimation for MoE architectures  
(see e.g. \cite{pavlitska2025moeuncertainty,zhang2023mofme}).
In this work, we focus on ensemble MoE architectures, as uncertainty quantification is more directly applicable for decision making and interpretability. 
In our method, we view the experts of the MoE model as an ensemble of models that can be used to extract both aleatoric and epistemic uncertainties.

\textbf{Uncertainty Estimation for Regression Tasks.} Deep learning regression models are increasingly required not only to provide accurate point estimates but also to quantify predictive uncertainty. A large body of research has focused on Bayesian neural networks, which place distributions over weights and approximate posterior inference using variational methods or Monte Carlo dropout, thereby producing predictive intervals \citep{gal2016dropout}. Another line of work employs ensembles of neural networks to capture both aleatoric and epistemic uncertainties, with randomized initialization or bootstrapped training providing diverse predictions \citep{lakshminarayanan2017simple}. More recently, post-hoc calibration techniques have been proposed, adapting classification-oriented approaches such as temperature scaling to regression settings, for instance by optimizing proper scoring rules or variance scaling factors \citep{kuleshov2018accurate}. Beyond probabilistic calibration, conformal prediction (CP) methods have gained attention due to their finite-sample coverage guarantees under minimal distributional assumptions. CP can be applied to regression to produce instance-dependent prediction intervals with guaranteed coverage, and has been extended to handle asymmetric intervals, distribution shift, and multi-target regression \citep{vovk2005conformal,cqr2019,Nizhar2025}. 

\textbf{Time Series Forecasting and Uncertainty Estimation.} Time series forecasting is a critical discipline in machine learning and statistics, focusing on predicting future values from a sequence of historical data points ordered by time. This field has wide-ranging applications, including financial market analysis, energy consumption forecasting, weather prediction, and medical prognosis. Traditional statistical methods, such as Autoregressive Integrated Moving Average (ARIMA) and Exponential Smoothing, have been foundational. However, their effectiveness is often limited by their assumption of linearity and their inability to capture complex, non-linear dependencies. More recently, deep learning models, employing  Transformers \citep{Yuqietal-2023-PatchTST,autoformer,reformer}, Multi-Layer Perceptrons (MLPs) \citep{wang2024timemixer, dlinear}, and Convolutional Neural Networks (CNNs) \citep{wu2023timesnet}, were shown to be effective in modeling temporal dynamics and long-range dependencies \citep{wang2024deep,time_series_survey,xwang2024deep}. The ability to quantify the uncertainty of a forecast, rather than providing just a single point estimate, is of paramount importance.  Uncertainty quantification provides a confidence interval for the prediction, which is crucial for risk management and informed decision-making. Some recent works have introduced uncertainty estimation to time series forecasting (see e.g. \cite{cini2025corel, wu2025eci}).
Given its wide-ranging applications, the importance of reporting uncertainty, and its challenging nature, time series forecasting serves as a highly suitable domain to evaluate the performance of MoGU.

\section{Uncertainty-based Mixture Model}\label{sec:method}
In this section, we introduce our uncertainty-based gating for the MoE framework. We begin by outlining the general formulation of MoE in Section~\ref{subsec:moe} and its extension to Mixture of Gaussians Experts (MoGE) in Section~\ref{subsec:mog}. Subsequently, we present our proposed method, MoGU, which extends the MoGE formulation to an uncertainty-based gating model (Section~\ref{subsec:mogu}). Finally, in Section~\ref{subsec:tsf_mogu}, we apply this mechanism to the task of time series forecasting. 
\subsection{The MoE Framework}\label{subsec:moe}
A general formulation for an MoE network \citep{jacobs1991adaptive} can be defined as follows:
\begin{equation}\label{eq:moe}
    x  \rightarrow  ( w_i(x), y_i(x)), \hspace{1cm} i=1,...,k 
\end{equation}
where $x$ denotes the input, $y_i$ is the prediction of the $i$-th expert and $w_i$ is the weight the model assigns to that expert's prediction. The model's output is then calculated as the weighted sum of these expert predictions:
\begin{equation}\label{eq:moe_pred}
    \hat{y} = \sum w_i(x){y_i}(x).
\end{equation}
Optimizing an MoE is achieved by minimizing the following loss:
\begin{equation}\label{eq:moe_loss}
    \mathcal{L}_{\scriptscriptstyle \text{MoE}}= \sum w_i(x)\mathcal{L}(y_i(x), y) 
\end{equation}
where $y$ is the ground truth label and $\mathcal{L}$ is the loss function for the target task. 

Typically, an MoE comprises a set of individual expert neural networks (often architecturally identical) that predict the outputs $y_i$, along with an additional gating neural module responsible for predicting the expert weights $w_i$. In its initial conception \citep{jacobs1991adaptive}, both the experts and the gating module were realized as feedforward networks (the latter incorporating a softmax layer for weight prediction). However, the underlying formulation is adaptable, and subsequent research has introduced diverse architectural implementations. Additionally, MoEs have also been implemented as layers within larger models~\citep{shazeer2017outrageously}, which we refer to as 'latent MoEs'. 
%In this work, we focus on the general formulation provided in Eq.(~\ref{eq:mog}).
\subsection{From MoE to MoGE}\label{subsec:mog}
We can add to each expert an uncertainty component that indicates how much the expert is confident in its decision: 
\begin{equation}\label{eq:mog}
    x  \rightarrow  ( w_i(x), y_i(x), \sigma^2_i(x)), \hspace{1cm} i=1,...,k. 
\end{equation}
We can interpret $\sigma^2_i(x)$ as a variance term associated with the $i$-th expert. 
The experts' predictions and their variances can be jointly trained by replacing the individual expert loss $\mathcal{L}$ in Eq.~(\ref{eq:moe_loss}) with the Gaussian Negative Log Likelihood (NLL) loss, denoted by $\mathcal{L}_{\scriptscriptstyle \text{NLLG}}$:
\begin{equation}
    \mathcal{L}_{\scriptscriptstyle \text{MoGE}}= \sum w_i(x)\mathcal{L_{  \scriptscriptstyle \text{  NLLG}}}(y;y_i(x), \sigma_i^2(x))
    \label{eq:mog_loss}
\end{equation}
with:
\begin{equation}
    \mathcal{L}_{\scriptscriptstyle \text{NLLG}}(y; \mu, \sigma^2) = \frac{1}{2}(\log(\max(\sigma^2, \epsilon)) + \frac{ (\mu - y)^2 }{\max(\sigma^2, \epsilon)})
\end{equation}
where $\epsilon$ is used for stability.

Similarly to the MoE formulation (Eq. (\ref{eq:moe_loss})), the weights $w_i(x)$ are obtained through a softmax layer, which is computed by a separate gating module in addition to the experts given the input. 

This model thus assumes that the conditional distribution of the labels $y$ given $x$  is a mixture of Gaussians. Therefore, at the inference step, the model prediction is given by:
 \begin{equation}
 \hat{y} = E ( y|x) = \sum w_i(x)y_i(x).
 \end{equation}
The law of total variance implies that:
\begin{equation}
\Var(y|x)  = \underbrace{\sum w_i(x) \sigma_i^2(x)}_{\text{aleatoric uncertainty}} +  \underbrace{\sum w_i(x) (\hat{y}-y_i(x))^2}_{\text{epistemic uncertainty}}.
\label{var-mog}
\end{equation}
The first term of (\ref{var-mog}) can be viewed as the \textit{aleatoric uncertainty} and the second term is the \textit{epistemic uncertainty} (see e.g. \citep{gal2016dropout}). Here, we use the experts and an ensemble of regression models (instead of extracting the ensemble from the dropout mechanism).
\subsection{From MoGE to MoGU: Mixture-of-Gaussians with Uncertainty-based gating }\label{subsec:mogu}
We now describe our proposed framework, which extends MoGE with Uncertainty-based Gating (MoGU).
Once we add an uncertainty term for each expert, we can also interpret this term as the expert’s relevance to the prediction task for the given input signal. We can thus transform the expert confidence information into relevance weights, allowing us to replace the standard input-based MoE gating mechanism, with a decision function that is based on expert uncertainties.
We next present an alternative model, where the gating mechanism is based on using the variance of expert predictions as an uncertainty weight when combining the experts.

We can view each expert as an independently sampled noisy version of the true value $y$:  $y_i \sim \mathcal{N} ( y, \sigma_i^2(x))$. It can be easily verified that the maximum likelihood estimation of $y$ based on the experts' decisions $y_1,...,y_k$ is: 
\begin{equation}\hat{y} = 
%\arg \max_y p ( y_1,...,y_k; y , \sigma_1, ..., \sigma_k ) = 
\arg \max_y \sum_i \log \mathcal{N} ( y_i, ; y , \sigma_i^2) = 
\sum_i w_i y_i
\end{equation}
 s.t.
\begin{equation}
    w_i = \frac{ \sigma_i^{-2} }{ \sum_j \sigma_j^{-2}}.
    \label{widef}
\end{equation}
In other words, each expert is weighted in inverse proportion to its variance (i.e., proportional to its precision).
In contrast to traditional MoEs where gating is learned as an auxiliary neural module, MoGU derives gating weights directly from uncertainty estimates, reframing expert selection as probabilistic inference rather than an additional prediction task.
We can thus substitute Eq. (\ref{widef})  in Eq.~(\ref{eq:mog_loss}), to obtain the following loss function:
\begin{equation}
    \mathcal{L}_{\scriptscriptstyle \text{MoGU}}= \sum_i  \frac{ \sigma_i^{-2}(x)}{\sum_j \sigma_j^{-2}(x)}\mathcal{L}_{\scriptscriptstyle \text{NLLG}}(y;y_i(x), \sigma_i^2(x)).
\end{equation}

%At the inference step, the model prediction is  % \begin{equation}
 %\hat{y} = E ( y|x) = \sum_i \frac{y_i(x)}{\sigma_i^2(x)}   /  \sum_i \frac{1}{\sigma_i^2(x)}
% \end{equation}
 Further substituting (\ref{widef}) in (\ref{var-mog}) we obtain the variance reported by the MoGU model:
\begin{equation}
\Var(y|x)  = \underbrace{ \frac{k}{\sum_j \sigma_j^{-2}(x)} }_{\text{aleatoric uncertainty}} +  \underbrace{\sum_i \frac{ \sigma_i^{-2}(x)}{\sum_j \sigma_j^{-2}(x)} (\hat{y}-y_i(x))^2}_{\text{epistemic uncertainty}}.
\label{eq:var-mogu}
\end{equation}

Note that here the aleatoric uncertainty (the first additive term of  (\ref{eq:var-mogu}))  is simply the harmonic mean of the variances of the individual expert predictions.

\subsection{Time Series Forecasting with MoGU}\label{subsec:tsf_mogu}
We demonstrate the application of the MoGU approach to multivariate time series forecasting. The forecasting task is to predict future values of a system with multiple interacting variables. Given a sequence of $T$ observations for $V$ variables, represented by the matrix $x \in \mathbb{R}^{T \times V}$, the objective is to forecast the future values
$y\in \mathbb{R}^{ h\times V}$ where $h$ is the forecasting horizon.

Traditional neural forecasting models (forecasting 'experts') typically follow a two-step process. First, a neural module $g$, such as a Multi-Layer Perceptron (MLP) or a Transformer, encodes the input time series $x$ into a latent representation. Second, a fully connected layer $f$ regresses the future values $y$ from the latent representation $g(x)$. This process can be generally expressed as:
\begin{equation}\label{eq:mog-tsf}
    x  \rightarrow  f(g(x)).
\end{equation}

To apply MoGU for time series forecasting, we need to extend  forecasting experts with an uncertainty component as described in Eq. (\ref{eq:mog}), by estimating the variance of the forecast in addition to the predicted values. 

We implement this extension by introducing an \textit{uncertainty head}, $f'$, which predicts the variance $\sigma^2$ from the latent representation $g(x)$. We parameterize $f'$ as an MLP with a single hidden layer matching the dimensions of $g(x)$. The output of this layer is then passed through a Softplus function to ensure the variance is always non-negative and to promote numerical stability during training:
\begin{equation}\label{eq:mog-tsf2}
    \sigma^2(x) = \log(1 + e^{f'(g(x))}).
\end{equation}
We estimate the uncertainty at the same resolution as the prediction; that is, the model estimates uncertainty per-variable, per-time step.

The complete MoGU forecasting process is given by the following equation:
\begin{equation}\label{eq:mog-tsf3}
    x  \rightarrow  ( w_i, f_i(g_i(x)), \sigma_i^2(x)), \hspace{1cm} i=1,...,k 
\end{equation}
where $w_i$ is computed as in Eq.~(\ref{widef})
and $\sigma_i^2(x)$ is defined in Eq. (\ref{eq:mog-tsf2}). The final forecasting prediction is the weighted combination of expert means.

We provide a pseudo-code for MoGU in our Appendix as well as a complete PyTorch implementation to reproduce the results reported in our paper.

%subsection{Uncertainty Calibration}\label{subsec:calib}   Conformal Prediction (CP) \cite{vovk2005conformal,angelopoulos2021gentle,zhou2025conformal} is %a general non-parametric confidence calibration method  which,  given a confidence value,  aims to build a confidence interval such that the probability that the correct value is within this set is at least the given value. One standard method for obtaining an instance-based calibrated interval is the Conformalized Quantile Regression (CQR) algorithm \cite{cqr2019}. A recent study showed that applying the CP scale to the prediction variance  \cite{Nizhar2025} yields state-of-the-art results.  Adding a variance term to the mixture model   enables applying CP to the variance to obtain an instance-based calibrated confidence interval. Given a labeled calibration set $(x_1,y_1),...,(x_n,y_n)$,  define the conformity scores: $s_i = |y_i-\hat{y}_i|/\sqrt{ \Var(y_i|x_i)}$. Let  $q$ be the $1-\alpha$ quantile of the scores. The CP theorem  \cite{vovk2005conformal} provides a coverage guarantee for the case of IID samples:  \begin{equation}  p (  y \in [ \hat{y}(x) - q \sigma , \hat{y}(x) + q\sigma] )  \ge 1-\alpha
% \end{equation}   s.t.  $\sigma^2 =  \Var(y|x)$ and $(x,y)$ is a labeled test point.  In our case of a time series, the IID assumption doesn't hold.   We empirically demonstrate that the coverage requirement remains satisfied.

\section{Experiments}\label{sec:experiments}
We evaluate the MoGU framework across a diverse suite of multivariate time series forecasting benchmarks to validate its effectiveness in both predictive accuracy and uncertainty quantification. Our evaluation is structured around three core objectives: (i) assessing MoGU's forecasting performance against state-of-the-art deterministic and probabilistic baselines, (ii) analyzing the fidelity and calibration of its uncertainty estimates, and (iii) investigating the robustness of the framework through extensive ablation studies. Section \ref{sec:experimental_setup} details the datasets, expert backbones, and implementation protocols. In Section \ref{subsec:tsf_results}, we present the main forecasting results, including a multi-seed analysis of training stability. Section \ref{subsec:unc_analysis} provides an in-depth investigation into uncertainty fidelity, focusing on conformal calibration and error-variance correlation. Finally, Section \ref{subsec:ablations} evaluates our key design choices, including gating logic, head architecture, temporal resolution, and loss function.

\subsection{Experimental setup}
\label{sec:experimental_setup}
\textbf{Datasets.}
We evaluate our method on eight widely used time series forecasting datasets~\citep{autoformer}: four Electricity Transformer Temperature (ETT) datasets (ETTh1, ETTh2, ETTm1, ETTm2)~\citep{informer}, as well as Electricity\footnote{
https://archive.ics.uci.edu/ml/datasets\\/ElectricityLoadDiagrams20112014}, Weather\footnote{
https://www.bgc-jena.mpg.de/wetter/}, Exchange~
\citep{lai2018modeling}, and Illness (ILI)\footnote{
https://gis.cdc.gov/grasp/fluview/fluportaldashboard.html}.
%\jacob{do we need to add references to the datasets? YS: added}

\textbf{Experimental Protocol.} Our experiments follow the standard protocol used in recent time series forecasting literature \citep{Yuqietal-2023-PatchTST,liu2023itransformer, wang2024deep}. For the ILI dataset, we use a forecast horizon length  $h \in \{24, 36, 48, 60\}$. For all other datasets, the forecast horizon length is selected from {96,192,336,720}. A look-back window of 96 is used for all experiments. We report performance using the Mean Absolute Error (MAE) and Mean Squared Error (MSE). 
The quality of uncertainty quantification is assessed via calibration and error-variance correlation analyses. We provide implementation details of our calibration analysis in the Appendix, Section~\ref{sec:appendix_calibration_impl}. For the variance analysis, compute the Pearson and Spearman correlation with respect to the prediction error. Specifically, for each individual variable, we correlate the model's reported uncertainty values with the corresponding MAE across all time points. We then average these correlation coefficients to get an overall measure.

\textbf{Expert Architecture.} MoGU is a general MoE framework compatible with various expert architectures. We evaluate it using three commonly benchmarked state-of-the-art expert models: iTransformer \citep{liu2023itransformer}, PatchTST \citep{Yuqietal-2023-PatchTST}, and DLinear \citep{dlinear}. These models represent different architectural approaches, including Transformer and MLP-based designs. 
\linespread{1.00}
\begin{table*}[t]
\centering
\scriptsize
\caption{Multivariate forecasting results (MSE) for a 96-step horizon. We compare a single expert against standard Mixture of Experts (MoE) and our Uncertainty-based gating (MoGU) across varying expert counts. Bold indicates the best result per dataset.}
\label{tab:main_results_num_experts}
\renewcommand{\arraystretch}{1.2}
\setlength{\tabcolsep}{6pt}
\begin{tabular}{l c cccc cccc}
\toprule
\textbf{Dataset} & \textbf{Single Expert} & \multicolumn{4}{c}{\textbf{MoE (Baseline)}} & \multicolumn{4}{c}{\textbf{MoGU (Ours)}} \\
\cmidrule(lr){2-2} \cmidrule(lr){3-6} \cmidrule(lr){7-10}
\textbf{Num. Experts} & 1 & 2 & 3 & 4 & 5 & 2 & 3 & 4 & 5 \\
\midrule
ETTh1 & 0.398 & 0.391 & 0.393 & 0.398 & 0.392 & 0.385 & \textbf{0.380} & 0.382 & 0.381 \\
ETTh2 & 0.295 & 0.307 & 0.299 & 0.305 & 0.311 & 0.284 & \textbf{0.283} & 0.286 & 0.286 \\
ETTm1 & 0.341 & 0.349 & 0.332 & 0.347 & 0.339 & 0.320 & 0.320 & 0.314 & \textbf{0.312} \\
ETTm2 & 0.188 & 0.186 & 0.179 & 0.180 & 0.177 & 0.179 & 0.179 & 0.176 & \textbf{0.175} \\
\bottomrule
\end{tabular}
\end{table*}
\begin{table*}[!h]
\centering
\scriptsize
\caption{Multivariate long-term forecasting results for MoE and MoGU. Prediction Horizons are $\{24, 36, 48, 60\}$ for ILI and $\{96, 192, 336, 720\}$ for others. Bold indicates superior performance within each expert architecture group (iTransformer or PatchTST).}
\label{tab:main_results}
\renewcommand{\arraystretch}{1.15}
\setlength{\tabcolsep}{8pt} % Increased spread for better clarity
\begin{tabular}{ll cccc cccc}
\toprule
\multicolumn{2}{c}{\textbf{Expert}} & \multicolumn{4}{c}{\textbf{iTransformer}} & \multicolumn{4}{c}{\textbf{PatchTST}} \\
\cmidrule(lr){3-6} \cmidrule(lr){7-10}
\multicolumn{2}{c}{\textbf{Mixture Type}} & \multicolumn{2}{c}{MoE} & \multicolumn{2}{c}{MoGU (ours)} & \multicolumn{2}{c}{MoE} & \multicolumn{2}{c}{MoGU (ours)} \\
\cmidrule(lr){3-4} \cmidrule(lr){5-6} \cmidrule(lr){7-8} \cmidrule(lr){9-10}
\textbf{Dataset} & \textbf{Horizon} & MAE & MSE & MAE & MSE & MAE & MSE & MAE & MSE \\
\midrule
\multirow{4}{*}{ETTh1} 
& 96  & 0.410 & 0.393 & \textbf{0.400} & \textbf{0.380} & \textbf{0.406} & \textbf{0.386} & 0.415 & 0.409 \\
& 192 & 0.432 & 0.437 & \textbf{0.431} & \textbf{0.436} & 0.448 & 0.459 & \textbf{0.443} & \textbf{0.453} \\
& 336 & 0.472 & 0.504 & \textbf{0.454} & \textbf{0.479} & 0.465 & 0.485 & \textbf{0.459} & \textbf{0.484} \\
& 720 & \textbf{0.489} & \textbf{0.500} & 0.491 & 0.501 & 0.494 & 0.510 & \textbf{0.483} & \textbf{0.485} \\
\midrule
\multirow{4}{*}{ETTh2} 
& 96  & 0.348 & 0.299 & \textbf{0.336} & \textbf{0.283} & 0.347 & 0.298 & \textbf{0.331} & \textbf{0.277} \\
& 192 & 0.396 & 0.377 & \textbf{0.387} & \textbf{0.361} & 0.400 & 0.375 & \textbf{0.386} & \textbf{0.357} \\
& 336 & 0.427 & \textbf{0.413} & \textbf{0.425} & 0.415 & 0.440 & 0.422 & \textbf{0.423} & \textbf{0.406} \\
& 720 & 0.447 & 0.435 & \textbf{0.442} & \textbf{0.421} & 0.460 & 0.443 & \textbf{0.447} & \textbf{0.426} \\
\midrule
\multirow{4}{*}{ETTm1} 
& 96  & 0.367 & 0.332 & \textbf{0.356} & \textbf{0.320} & 0.371 & 0.337 & \textbf{0.362} & \textbf{0.326} \\
& 192 & 0.396 & 0.382 & \textbf{0.379} & \textbf{0.363} & 0.398 & \textbf{0.380} & \textbf{0.393} & 0.389 \\
& 336 & 0.411 & 0.407 & \textbf{0.404} & \textbf{0.400} & \textbf{0.407} & \textbf{0.400} & \textbf{0.407} & \textbf{0.400} \\
& 720 & 0.460 & 0.500 & \textbf{0.438} & \textbf{0.466} & 0.448 & 0.465 & \textbf{0.442} & \textbf{0.460} \\
\midrule
\multirow{4}{*}{ETTm2} 
& 96  & 0.261 & \textbf{0.179} & \textbf{0.260} & \textbf{0.179} & 0.264 & 0.177 & \textbf{0.259} & \textbf{0.175} \\
& 192 & 0.306 & 0.246 & \textbf{0.302} & \textbf{0.245} & 0.308 & 0.247 & \textbf{0.303} & \textbf{0.242} \\
& 336 & 0.345 & 0.307 & \textbf{0.339} & \textbf{0.301} & \textbf{0.346} & \textbf{0.304} & \textbf{0.346} & 0.307 \\
& 720 & 0.401 & 0.403 & \textbf{0.395} & \textbf{0.397} & 0.405 & 0.408 & \textbf{0.403} & \textbf{0.405} \\
\midrule
\multirow{4}{*}{ILI} 
& 24 & 0.864 & 1.786 & \textbf{0.827} & \textbf{1.756} & 0.866 & 1.871 & \textbf{0.822} & \textbf{1.848} \\
& 36 & 0.882 & 1.746 & \textbf{0.825} & \textbf{1.629} & 0.875 & 1.875 & \textbf{0.835} & \textbf{1.801} \\
& 48 & 0.948 & 1.912 & \textbf{0.843} & \textbf{1.634} & 0.878 & \textbf{1.798} & \textbf{0.844} & 1.818 \\
& 60 & 0.979 & 1.986 & \textbf{0.881} & \textbf{1.692} & 0.904 & 1.864 & \textbf{0.864} & \textbf{1.831} \\
\midrule
\multirow{4}{*}{Weather} 
& 96  & 0.253 & 0.208 & \textbf{0.249} & \textbf{0.207} & 0.237 & 0.196 & \textbf{0.230} & \textbf{0.188} \\
& 192 & \textbf{0.283} & \textbf{0.246} & \textbf{0.283} & 0.251 & 0.268 & 0.235 & \textbf{0.265} & \textbf{0.232} \\
& 336 & \textbf{0.315} & \textbf{0.296} & 0.317 & 0.300 & 0.308 & 0.291 & \textbf{0.303} & \textbf{0.287} \\
& 720 & \textbf{0.361} & \textbf{0.369} & \textbf{0.361} & 0.371 & 0.353 & 0.363 & \textbf{0.351} & \textbf{0.361} \\
\midrule
\multirow{4}{*}{Electricity} 
& 96  & \textbf{0.235} & \textbf{0.144} & 0.238 & 0.148 & \textbf{0.248} & \textbf{0.161} & 0.257 & 0.169 \\
& 192 & 0.254 & \textbf{0.162} & \textbf{0.251} & 0.163 & \textbf{0.258} & \textbf{0.170} & 0.263 & 0.179 \\
& 336 & \textbf{0.269} & \textbf{0.175} & \textbf{0.269} & 0.179 & \textbf{0.276} & \textbf{0.188} & 0.286 & 0.200 \\
& 720 & \textbf{0.297} & \textbf{0.204} & 0.302 & 0.216 & \textbf{0.314} & \textbf{0.231} & 0.319 & 0.242 \\
\midrule
\multicolumn{2}{l}{\textbf{Num. Wins}} & 4 & 9 & \textbf{21} & \textbf{18} & 5 & 8 & \textbf{21} & \textbf{19} \\
\bottomrule
\end{tabular}
\end{table*}
\begin{table*}[h!]
\centering
\scriptsize
\caption{Multivariate forecasting results (MSE/MAE) for a 96-step horizon across three expert architectures: DLinear, iTransformer, and PatchTST. We compare standard MoE against MoGU (ours). Bold indicates the superior performance within each architecture group.}
\label{tab:main_results_exchange}
\renewcommand{\arraystretch}{1.2}
\setlength{\tabcolsep}{5pt}
\begin{tabular}{l cc cc cc cc cc cc}
\toprule
\textbf{Expert} & \multicolumn{4}{c}{\textbf{DLinear}} & \multicolumn{4}{c}{\textbf{iTransformer}} & \multicolumn{4}{c}{\textbf{PatchTST}} \\
\cmidrule(lr){2-5} \cmidrule(lr){6-9} \cmidrule(lr){10-13}
\textbf{Mixture Type} & \multicolumn{2}{c}{MoE} & \multicolumn{2}{c}{MoGU (ours)} & \multicolumn{2}{c}{MoE} & \multicolumn{2}{c}{MoGU (ours)} & \multicolumn{2}{c}{MoE} & \multicolumn{2}{c}{MoGU (ours)} \\
\cmidrule(lr){2-3} \cmidrule(lr){4-5} \cmidrule(lr){6-7} \cmidrule(lr){8-9} \cmidrule(lr){10-11} \cmidrule(lr){12-13}
\textbf{Dataset} & MAE & MSE & MAE & MSE & MAE & MSE & MAE & MSE & MAE & MSE & MAE & MSE \\
\midrule
Exchange & 0.213 & 0.086 & \textbf{0.209} & \textbf{0.080} & 0.218 & 0.096 & \textbf{0.208} & \textbf{0.089} & \textbf{0.201} & 0.086 & 0.202 & \textbf{0.084} \\
ETTh1    & \textbf{0.400} & \textbf{0.382} & \textbf{0.400} & \textbf{0.382} & 0.410 & 0.393 & \textbf{0.400} & \textbf{0.380} & \textbf{0.406} & \textbf{0.386} & 0.415 & 0.409 \\
ETTh2    & 0.373 & 0.320 & \textbf{0.366} & \textbf{0.308} & 0.348 & 0.299 & \textbf{0.336} & \textbf{0.283} & 0.347 & 0.298 & \textbf{0.331} & \textbf{0.277} \\
ETTm1    & \textbf{0.360} & \textbf{0.322} & 0.363 & 0.338 & 0.367 & 0.332 & \textbf{0.356} & \textbf{0.320} & 0.371 & 0.337 & \textbf{0.362} & \textbf{0.326} \\
ETTm2    & 0.285 & 0.189 & \textbf{0.271} & \textbf{0.183} & 0.261 & \textbf{0.179} & \textbf{0.260} & \textbf{0.179} & 0.264 & 0.177 & \textbf{0.259} & \textbf{0.175} \\
\bottomrule
\end{tabular}
\end{table*}
\begin{table}[t]
\centering
\scriptsize
\caption{Robustness analysis: Mean $\pm$ Standard Deviation across five seeds. Results are reported for a 96-step forecast horizon using iTransformer as the expert backbone. MoGU consistently exhibits lower variance and superior mean accuracy compared to standard MoE.}
\label{tab:prediction_variance}
\renewcommand{\arraystretch}{1.2}
\setlength{\tabcolsep}{3pt}
\begin{tabular}{l cc cc}
\toprule
\textbf{Dataset} & \multicolumn{2}{c}{\textbf{MoE (Baseline)}} & \multicolumn{2}{c}{\textbf{MoGU (Ours)}} \\
\cmidrule(lr){2-3} \cmidrule(lr){4-5}
\textbf{Metric} & MAE & MSE & MAE & MSE \\
\midrule
ETTh1 & 0.4093 $\pm$ 0.0024 & 0.3945 $\pm$ 0.0043 & \textbf{0.4006 $\pm$ 0.0010} & \textbf{0.3815 $\pm$ 0.0016} \\
ETTh2 & 0.3502 $\pm$ 0.0046 & 0.3048 $\pm$ 0.0094 & \textbf{0.3384 $\pm$ 0.0018} & \textbf{0.2853 $\pm$ 0.0020} \\
ETTm1 & 0.3699 $\pm$ 0.0029 & 0.3380 $\pm$ 0.0075 & \textbf{0.3535 $\pm$ 0.0019} & \textbf{0.3167 $\pm$ 0.0024} \\
ETTm2 & 0.2618 $\pm$ 0.0019 & 0.1787 $\pm$ 0.0024 & \textbf{0.2560 $\pm$ 0.0022} & \textbf{0.1741 $\pm$ 0.0017} \\
\bottomrule
\end{tabular}
\end{table}
\begin{table*}[t]
\centering
\scriptsize
\caption{Efficiency of predictive intervals (Mean $\pm$ Std) for a 0.90 target coverage.}
\label{tab:calibration_comparison}
\renewcommand{\arraystretch}{1.2}
\setlength{\tabcolsep}{8pt} % Slightly reduced to fit the extra column comfortably
\begin{tabular}{lll cc}
\toprule
\textbf{Dataset} & \textbf{Model} & \textbf{Calibration} & \textbf{Coverage} $\uparrow$ & \textbf{Avg. Width} $\downarrow$ \\
\midrule
\multirow{6}{*}{ETTh1} 
& \textbf{MoGU} & CPVS & 0.9073 $\pm$ 0.0009 & \textbf{1.8005 $\pm$ 0.0121}\\
& MoGE & CPVS & 0.9077 $\pm$ 0.0012 & 1.8128 $\pm$ 0.0203 \\
& Single Gaussian & CPVS & 0.9071 $\pm$ 0.0006 & 1.8423 $\pm$ 0.0149 \\
& MoE & CP-fixed & 0.9087 $\pm$ 0.0005 & 2.0005 $\pm$ 0.0125 \\
& Single Expert & CP-fixed & 0.9086 $\pm$ 0.0003 & 2.0098 $\pm$ 0.0128 \\
& Single Expert & CQR & 0.8811 $\pm$ 0.0098 & 3.7513 $\pm$ 0.2973\\
\midrule
\multirow{6}{*}{ETTm1} 
& \textbf{MoGU} & CPVS & 0.8982 $\pm$ 0.0002 & \textbf{1.5083 $\pm$ 0.0101}\\
& MoGE & CPVS & 0.8979 $\pm$ 0.0001 & 1.5485 $\pm$ 0.0092 \\
& Single Gaussian & CPVS & 0.8978 $\pm$ 0.0003 & 1.5792 $\pm$ 0.0106 \\
& MoE & CP-fixed & 0.8991 $\pm$ 0.0003 & 1.7151 $\pm$ 0.0234 \\
& Single Expert & CP-fixed & 0.8993 $\pm$ 0.0004 & 1.7180 $\pm$ 0.0261 \\
& Single Expert & CQR & 0.8891 $\pm$ 0.0011 & 2.9467 $\pm$ 0.1967 \\
\bottomrule
\end{tabular}
\end{table*}
\begin{table}[h!]
\centering
\scriptsize
\caption{Pearson ($R$) and Spearman ($\rho$) correlation coefficients between MoGU's reported uncertainty and the MAE of its predictions (Horizon: 96). Results are averaged across variables; all correlations are statistically significant ($p < 10^{-5}$). Total uncertainty reflects the combination of Aleatoric (A) and Epistemic (E) components (Eq.~\ref{eq:var-mogu}).}
\label{tab:unc_corr}
\renewcommand{\arraystretch}{1.1}
\setlength{\tabcolsep}{6pt}
\begin{tabular}{ll cc cc cc}
\toprule
\multirow{2}{*}{\textbf{Backbone}} & \multirow{2}{*}{\textbf{Dataset}} & \multicolumn{2}{c}{\textbf{Aleatoric (A)}} & \multicolumn{2}{c}{\textbf{Epistemic (E)}} & \multicolumn{2}{c}{\textbf{Total (A+E)}} \\
\cmidrule(lr){3-4} \cmidrule(lr){5-6} \cmidrule(lr){7-8}
& & $R$ & $\rho$ & $R$ & $\rho$ & $R$ & $\rho$ \\
\midrule
\multirow{4}{*}{iTransformer} 
& ETTh1 & 0.25 & 0.22 & 0.03 & 0.04 & 0.25 & 0.22 \\
& ETTh2 & 0.15 & 0.20 & 0.08 & 0.15 & 0.15 & 0.21 \\
& ETTm1 & 0.27 & 0.29 & 0.10 & 0.13 & 0.27 & 0.30 \\
& ETTm2 & 0.15 & 0.17 & 0.13 & 0.24 & 0.16 & 0.19 \\
\midrule
\multirow{4}{*}{PatchTST} 
& ETTh1 & 0.26 & 0.23 & 0.05 & 0.05 & 0.26 & 0.23 \\
& ETTh2 & 0.14 & 0.17 & 0.12 & 0.20 & 0.14 & 0.17 \\
& ETTm1 & 0.31 & 0.30 & 0.07 & 0.11 & 0.31 & 0.30 \\
& ETTm2 & 0.11 & 0.11 & 0.14 & 0.25 & 0.11 & 0.11 \\
\bottomrule
\end{tabular}
\end{table}
\begin{table}[h!]
\centering
\scriptsize
\caption{Ablation study of gating mechanisms (Horizon: 96). We compare MoE, MoGE, and our uncertainty-driven gating (MoGU). Bold indicates the best result for each backbone and dataset.}
\label{tab:ablations_gating}
\renewcommand{\arraystretch}{1.1}
\setlength{\tabcolsep}{5pt}
\begin{tabular}{ll cc cc cc}
\toprule
\multirow{2}{*}{\textbf{Backbone}} & \multirow{2}{*}{\textbf{Dataset}} & \multicolumn{2}{c}{\textbf{MoE}} & \multicolumn{2}{c}{\textbf{MoGE}} & \multicolumn{2}{c}{\textbf{MoGU (Ours)}} \\
\cmidrule(lr){3-4} \cmidrule(lr){5-6} \cmidrule(lr){7-8}
& & MAE & MSE & MAE & MSE & MAE & MSE \\
\midrule
\multirow{4}{*}{iTransformer} 
& ETTh1 & 0.410 & 0.393 & 0.403 & 0.387 & \textbf{0.400} & \textbf{0.380} \\
& ETTh2 & 0.348 & 0.299 & 0.340 & 0.288 & \textbf{0.336} & \textbf{0.283} \\
& ETTm1 & 0.367 & 0.332 & 0.360 & 0.326 & \textbf{0.356} & \textbf{0.320} \\
& ETTm2 & 0.261 & 0.179 & \textbf{0.256} & \textbf{0.175} & 0.260 & 0.179 \\
\midrule
\multirow{4}{*}{PatchTST} 
& ETTh1 & \textbf{0.406} & \textbf{0.386} & 0.420 & 0.413 & 0.415 & 0.409 \\
& ETTh2 & 0.347 & 0.298 & 0.343 & 0.291 & \textbf{0.331} & \textbf{0.277} \\
& ETTm1 & 0.371 & 0.337 & 0.372 & 0.337 & \textbf{0.362} & \textbf{0.326} \\
& ETTm2 & 0.264 & 0.177 & \textbf{0.259} & 0.176 & \textbf{0.259} & \textbf{0.175} \\
\bottomrule
\end{tabular}
\end{table}

\begin{table}[h!]
\centering
\scriptsize
\caption{Ablation study of the uncertainty head architecture (Horizon: 96). We compare a single Fully Connected (FC) layer against a Multi-Layer Perceptron (MLP). Bold indicates the best result per dataset.}
\label{tab:ablations_head_arc}
\renewcommand{\arraystretch}{1.1}
\setlength{\tabcolsep}{10pt}
\begin{tabular}{ll cc cc}
\toprule
\multirow{2}{*}{\textbf{Backbone}} & \multirow{2}{*}{\textbf{Dataset}} & \multicolumn{2}{c}{\textbf{FC Head}} & \multicolumn{2}{c}{\textbf{MLP Head}} \\
\cmidrule(lr){3-4} \cmidrule(lr){5-6}
& & MAE & MSE & MAE & MSE \\
\midrule
\multirow{4}{*}{iTransformer} 
& ETTh1 & \textbf{0.399} & 0.383 & 0.400 & \textbf{0.380} \\
& ETTh2 & 0.338 & 0.286 & \textbf{0.336} & \textbf{0.283} \\
& ETTm1 & 0.357 & 0.321 & \textbf{0.356} & \textbf{0.320} \\
& ETTm2 & 0.261 & 0.178 & \textbf{0.260} & \textbf{0.179} \\
\midrule
\multirow{4}{*}{PatchTST} 
& ETTh1 & \textbf{0.410} & 0.401 & 0.415 & \textbf{0.409} \\
& ETTh2 & 0.340 & 0.285 & \textbf{0.331} & \textbf{0.277} \\
& ETTm1 & \textbf{0.356} & \textbf{0.320} & 0.362 & 0.326 \\
& ETTm2 & 0.260 & \textbf{0.174} & \textbf{0.259} & 0.175 \\
\bottomrule
\end{tabular}
\end{table}
{\bf Implementation and Training Details.}
We implemented MoGU in PyTorch~\citep{paszke2019pytorch}. For the expert architecture, we extended the existing implementations of PatchTST, iTransformer, and DLinear available from the Time Series Library (TSLib)~\citep{wang2024deep}, to incorporate uncertainty estimation as detailed in Section \ref{subsec:tsf_mogu}. Following the standard configuration provided by TSLib for training time series forecasting architectures, we trained all models using the Adam optimizer for a maximum of 10 epochs, with early stopping patience set to 3 epochs. The learning rate was set to $\lambda=0.001$ for the Weather and Electricity datasets, and $\lambda=0.0001$ for all other datasets. All experiments were conducted on a single NVIDIA A100 80GB GPU.

\subsection{Results}\label{sec:results}
\subsubsection{Time Series Forecasting with MoGU}\label{subsec:tsf_results}
Table \ref{tab:main_results_num_experts} compares MoGU's performance against single-expert and standard MoE configurations on the ETT datasets. Utilizing iTransformer as the expert backbone and scaling the number of experts from 2 to 5, MoGU consistently achieves superior predictive accuracy compared to both baselines.

Tables~\ref{tab:main_results} and \ref{tab:main_results_exchange} provide a comprehensive comparison between a three-expert MoE and MoGU across a variety of multivariate forecasting datasets and horizon lengths. MoGU outperforms standard MoE in the majority of benchmarks using iTransformer, PatchTST, and DLinear as expert architectures. 

To ensure statistical reliability and assess the stability of our proposed gating mechanism, we report the mean and standard deviation for both MoE and MoGU across five independent random initializations. These results, summarized in Table~\ref{tab:prediction_variance}, demonstrate that MoGU not only improves predictive accuracy but also reduces performance variance compared to standard MoE configurations. 

\subsubsection{Uncertainty Estimation and Calibration Analysis}\label{subsec:unc_analysis}
We evaluate the fidelity of the uncertainty quantification produced by MoGU. We focus on two primary dimensions: the calibration of predictive intervals via conformal prediction and the statistical correlation between reported uncertainty and empirical error.

\textbf{Confidence Interval Calibration using Conformal Prediction.} 
MoGU’s routing is uncertainty-driven, so its uncertainty estimates are not only auxiliary outputs but central to the model. We validate their usefulness through conformal prediction (CP) \cite{vovk2005conformal}.  Improved uncertainty estimation should translate into tighter intervals while preserving coverage.
We conduct a comparative calibration analysis by implementing three CP frameworks to generate valid prediction intervals. The first, denoted by CP-fixed, is based on the 
conformity score $|y-\hat{y}|$. This method computes a fixed-size confidence interval. The second method is CQR \cite{cqr2019}, which is the standard method for obtaining an instance-based calibrated interval. 
CQR requires training a separate quantile regression network to predict the conformity score quantile. 
The third method, denoted by CP Variance Scaling (CPVS), is based on the conformity score $|y-\hat{y}|/\sigma(y|x)$ where $\sigma^2(y|x)$ is the sample-based estimated prediction variance. 
CPVS is relevant in cases where the model outputs the prediction variance.
Time-series forecasting fundamentally violates the CP exchangeability assumption due to non-stationarity and continuous temporal distribution shifts. Consequently, applying a static calibration set, derived from past data, often results in prediction intervals that are either invalid (under-coverage) or unnecessarily conservative (over-wide) as the data dynamics evolve.
The domain shift can be partially mitigated using a rolling window mechanism where the calibration set is dynamically updated at each time step \cite{gibbs2021adaptive}.

%Table \ref{tab:calibration_comparison} presents the calibration results, showing that the average interval length (Avg. Width) obtained by MoGU is smaller than those achieved by MoE and MoG. Although this is not the focus of this study, we also observe that CQR yields the worst results, likely due to the non-robustness of the pinball loss used for training.
Table \ref{tab:calibration_comparison} summarizes the calibration results, showing that the average interval width (Avg. Width) obtained by MoGU is lower than those achieved by MoE and MoGE. Notably, CQR yields the least efficient results, likely due to the non-robustness of the pinball loss used for training.

\textbf{Variance Analysis}
  \begin{figure}[h!]
  \centering
\includegraphics[width=7cm]{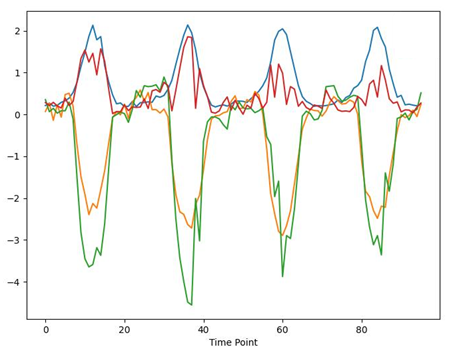}
\includegraphics[width=7cm]{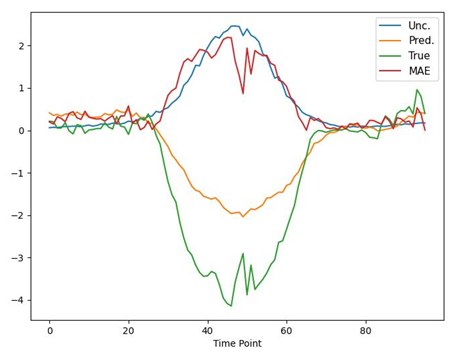}
\caption{Example forecasts along with the ground truth, the MAE and uncertainty reported by MoGU with three experts. The forecasts for the ETTh1 dataset (a) were generated using PatchTST as the expert architecture, while those for ETTm1 (b) were generated using iTransformer.}
        \label{fig:unc_graph}
\end{figure}
To assess how well MoGU's reported uncertainty aligns with its actual prediction errors, we further compute the Pearson (R) and Spearman ($\rho$) correlation coefficients between them. Table \ref{tab:unc_corr} presents these coefficients for the aleatoric, epistemic, and total uncertainties (as defined in Eq. \ref{eq:var-mogu}).

We observe a statistically significant positive correlation between MoGU's uncertainty estimates and the Mean Absolute Error (MAE) of its predictions. Interestingly, the correlation with aleatoric uncertainty is typically higher than with epistemic uncertainty. Since aleatoric uncertainty represents the inherent randomness in the data itself, this correlation suggests that the model can use uncertainty estimates to identify data points where irreducible randomness makes accurate predictions difficult, thereby leading to higher errors.

Fig. \ref{fig:unc_graph} illustrates the relationship between MoGU's prediction error and uncertainty estimates by showing the predicted and ground truth values alongside the MAE and reported uncertainty for representative examples. The uncertainty at each time point closely follows the prediction error. 
Appendix~\ref{sec:appendix_var_analysis} provides variable-wise correlation heatmaps (Fig.~\ref{fig:heatmaps}) and an analysis of per-expert variance and weight distributions. These results demonstrate that the positive error-variance correlation persists at the variable level (Fig.~\ref{fig:variance_error_correlation}) and that expert utilization remains balanced, with no observed expert collapse (Table~\ref{tab:expert_weights_distribution}).
%We further show the Pearson correlation heatmaps in Fig. \ref{fig:heatmaps} in our Appendix. These heatmaps further visualize the relationship between the Mean Absolute Error (MAE) of MoGU's predictions and its reported uncertainties (aleatoric, epistemic, and total), when using MoGU with three iTransformer experts. The analysis is presented per variable for each of the ETT datasets, highlighting the extent to which different uncertainty components correlate with predictive error. While the correlation between uncertainty and MAE varies among variables, it remains consistently positive.

%We further provide a granular analysis of per-expert variance, prediction error and weight distributions in Appendix~\ref{sec:appendix_var_analysis}. Our results demonstrate a positive correlation between predicted variance and actual error, also at the expert-level and confirm that MoGU effectively mitigates expert collapse. 
%We discuss limitations and future work in our Appendix, Section~\ref{sec:appendix_limitations_future}. 
\subsection{Ablations}\label{subsec:ablations}
We conducted an ablation study to evaluate our key design choices. For all experiments, we used a configuration with three experts.

\textbf{Gating Mechanism.} Table \ref{tab:ablations_gating} compares our MoGU to a standard input-based gating mechanism \citep{jacobs1991adaptive}, when employed by a deterministic MoE and with a MoGE. The input-based method utilizes a separate neural module to predict weights by processing the input before a softmax layer. We evaluated the MoE, MoGE and MoGU methods on four ETT datasets using iTransformer and PatchTST as the expert architectures. Our uncertainty-based gating consistently resulted in a lower prediction error.

\textbf{Uncertainty Head Architecture.}
We also evaluated the design of our uncertainty head, which is implemented as a shallow Multi-Layer Perceptron (MLP) with a single hidden fully connected layer. Table \ref{tab:ablations_head_arc} compares this to an alternative using only a single fully connected layer. The MLP alternative performed better in most cases, though the performance difference was relatively small.

\textbf{Resolution of Uncertainty Estimation.}
Table \ref{tab:ablations_time} in our Appendix explores an alternative where the expert estimates uncertainty at the variable level ('Time-Fixed'), rather than for each individual time point ('Time-Varying'). Predicting uncertainty at the higher resolution of a single time point yielded better results, demonstrating the advantage of our framework's ability to provide high-resolution uncertainty predictions. We note that our framework is flexible and supports both configurations.

Additional ablations for our \textbf{Loss Function} are provided in the Appendix (Section~\ref{subsec:app_ablations}).
\subsection{Limitations and Future Work}
\label{sec:limitations_future}
While MoGU shows promise for time series forecasting, broadening its scope to other regression (and classification) tasks will further validate its robustness and generalization. In addition, adapting its dense gating for sparse architectures like those in LLMs remains a challenge for future work.
\section{Conclusion}
We introduced MoGU, a novel extension of MoE for time series forecasting. Instead of using traditional input-based gating, MoGU's gating mechanism aggregates expert predictions based on their individual uncertainty (variance) estimates. This approach led to superior performance over single-expert and conventional MoE models across various benchmarks, architectures, and time horizons. Our results suggest a promising new direction for MoEs: integrating probabilistic information directly into the gating process for more robust and reliable models.

\bibliography{paper}

@inproceedings{cqr2019,
  title={Conformalized quantile regression},
  author={Romano, Yaniv and Patterson, Evan and Candes, Emmanuel},
  booktitle={NeurIPS},
  year={2019}
}

@book{vovk2005conformal,
  title={Algorithmic learning in a random world},
  author={Vovk, Vladimir and Gammerman, Alexander and Shafer, Glenn},
  volume={29},
  year={2005},
  publisher={Springer}
}

@inproceedings{lai2018modeling,
  title={Modeling long-and short-term temporal patterns with deep neural networks},
  author={Lai, Guokun and Chang, Wei-Cheng and Yang, Yiming and Liu, Hanxiao},
  booktitle={The 41st international ACM SIGIR conference on research \& development in information retrieval},
  pages={95--104},
  year={2018}
}

@inproceedings{Nizhar2025,
title={Clinical Measurements with Calibrated Instance-Dependent Confidence Interval},
Author ={ Rotem Nizhar and  Lior Frenkel and  Jacob Goldberger}, 
booktitle={Medical Imaging with Deep Learning (MIDL)},
year=2025
}

@inproceedings{fedus2022switch,
  title={Switch Transformers: Scaling to trillion parameter models with simple and efficient sparsity},
  author={Fedus, William and Zoph, Barret and Shazeer, Noam},
  booktitle={International Conference on Learning Representations (ICLR)},
  year={2022}
}

@inproceedings{shazeer2017outrageously,
  title={Outrageously large neural networks: The sparsely-gated mixture-of-experts layer},
  author={Shazeer, Noam and Mirhoseini, Azalia and Maziarz, Krzysztof and Davis, Andy and Le, Quoc V. and Hinton, Geoffrey and Dean, Jeffrey},
  booktitle={International Conference on Learning Representations (ICLR)},
  year={2017}
}

@article{time_series_survey,
  title={Time-series forecasting with deep learning: a survey},
  author={Lim, Bryan and Zohren, Stefan},
  journal={Philosophical Transactions of the Royal Society A},
  volume={379},
  number={2194},
  pages={20200209},
  year={2021},
  publisher={The Royal Society Publishing}
}

@inproceedings{gal2016dropout,
  title={Dropout as a Bayesian approximation: Representing model uncertainty in deep learning},
  author={Gal, Yarin and Ghahramani, Zoubin},
  booktitle={Proceedings of the 33rd International Conference on Machine Learning (ICML)},
  pages={1050--1059},
  year={2016}
}

@inproceedings{lakshminarayanan2017simple,
  title={Simple and scalable predictive uncertainty estimation using deep ensembles},
  author={Lakshminarayanan, Balaji and Pritzel, Alexander and Blundell, Charles},
  booktitle={Advances in Neural Information Processing Systems (NeurIPS)},
  pages={6402--6413},
  year={2017}
}

@inproceedings{kuleshov2018accurate,
  title={Accurate uncertainties for deep learning using calibrated regression},
  author={Kuleshov, Volodymyr and Fenner, Nathan and Ermon, Stefano},
  booktitle={Proceedings of the 35th International Conference on Machine Learning (ICML)},
  pages={2796--2804},
  year={2018}
}

@inproceedings{wu2023timesnet,
  title={TimesNet: Temporal 2D-Variation Modeling for General Time Series Analysis},
  author={Haixu Wu and Tengge Hu and Yong Liu and Hang Zhou and Jianmin Wang and Mingsheng Long},
  booktitle={International Conference on Learning Representations},
  year={2023},
}

@inproceedings{liu2023itransformer,
  title={iTransformer: Inverted Transformers Are Effective for Time Series Forecasting},
  author={Liu, Yong and Hu, Tengge and Zhang, Haoran and Wu, Haixu and Wang, Shiyu and Ma, Lintao and Long, Mingsheng},
  booktitle={The Twelfth International Conference on Learning Representations},
  year={2023}
}

@article{autoformer,
  title={Autoformer: Decomposition transformers with auto-correlation for long-term series forecasting},
  author={Wu, Haixu and Xu, Jiehui and Wang, Jianmin and Long, Mingsheng},
  journal={Advances in neural information processing systems},
    year={2021}
}

@inproceedings{dlinear,
  title={Are transformers effective for time series forecasting?},
  author={Zeng, Ailing and Chen, Muxi and Zhang, Lei and Xu, Qiang},
  booktitle={Proceedings of the AAAI conference on artificial intelligence},
  volume={37},
  number={9},
  pages={11121--11128},
  year={2023}
}

@article{reformer,
  title={Reformer: The efficient transformer},
  author={Kitaev, Nikita and Kaiser, {\L}ukasz and Levskaya, Anselm},
  journal={arXiv preprint arXiv:2001.04451},
  year={2020}
}

@inproceedings{informer,
  title={Informer: Beyond efficient transformer for long sequence time-series forecasting},
  author={Zhou, Haoyi and Zhang, Shanghang and Peng, Jieqi and Zhang, Shuai and Li, Jianxin and Xiong, Hui and Zhang, Wancai},
  booktitle={Proceedings of the AAAI conference on artificial intelligence},
  volume={35},
  number={12},
  pages={11106--11115},
  year={2021}
}

@article{wang2024timemixer,
  title={Timemixer: Decomposable multiscale mixing for time series forecasting},
  author={Wang, Shiyu and Wu, Haixu and Shi, Xiaoming and Hu, Tengge and Luo, Huakun and Ma, Lintao and Zhang, James Y and Zhou, Jun},
  journal={arXiv preprint arXiv:2405.14616},
  year={2024}
}

@article{wang2024deep,
  title={Deep Learning for Multivariate Time Series Imputation: A Survey},
  author={Wang, Jun and Du, Wenjie and Cao, Wei and Zhang, Keli and Wang, Wenjia and Liang, Yuxuan and Wen, Qingsong},
  journal={arXiv preprint arXiv:2402.04059},
  year={2024}
}

@article{xwang2024deep,
  title={Deep time series models: A comprehensive survey and benchmark},
  author={Wang, Yuxuan and Wu, Haixu and Dong, Jiaxiang and Liu, Yong and Long, Mingsheng and Wang, Jianmin},
  journal={arXiv preprint arXiv:2407.13278},
  year={2024}
}

@article{jacobs1991adaptive,
  title={Adaptive mixtures of local experts},
  author={Jacobs, Robert A and Jordan, Michael I and Nowlan, Steven J and Hinton, Geoffrey E},
  journal={Neural computation},
  volume={3},
  number={1},
  pages={79--87},
  year={1991},
  publisher={MIT Press}
}

@inproceedings{Yuqietal-2023-PatchTST,
  title     = {A Time Series is Worth 64 Words: Long-term Forecasting with Transformers},
  author    = {Nie, Yuqi and
               H. Nguyen, Nam and
               Sinthong, Phanwadee and 
               Kalagnanam, Jayant},
  booktitle = {International Conference on Learning Representations},
  year      = {2023}
}

@inproceedings{paszke2019pytorch,
  title={Pytorch: An imperative style, high-performance deep learning library},
  author={Paszke, Adam and Gross, Sam and Massa, Francisco and Lerer, Adam and Bradbury, James and Chanan, Gregory and Killeen, Trevor and Lin, Zeming and Gimelshein, Natalia and Antiga, Luca and others},
  booktitle={NeurIPS},
  year={2019}
}

@article{pavlitska2025moeuncertainty,
  title   = {Extracting Uncertainty Estimates from Mixtures of Experts for Semantic Segmentation},
  author  = {Pavlitska, Inna and Maillard, Adrien and Matejek, Brian and Lucic, Mario and Houlsby, Neil},
  journal = {arXiv preprint arXiv:2509.04816},
  year    = {2025}
}

@article{zhang2023mofme,
  title   = {Efficient Deweather Mixture-of-Experts with Uncertainty-aware Feature-wise Linear Modulation},
  author  = {Zhang, Yunbo and Chen, Yanhua and Li, Mingyang and Gao, Yujing and Zhang, Zhiqiang},
  journal = {arXiv preprint arXiv:2312.16610},
  year    = {2023}
}

@inproceedings{cini2025corel,
  title     = {Relational Conformal Prediction for Correlated Time Series},
  author    = {Cini, Andrea and Bogunovic, Ilija and Pavez, Eduardo and Cand{\`e}s, Emmanuel J.},
  booktitle = {Proceedings of the International Conference on Machine Learning (ICML)},
  year      = {2025},
  note      = {arXiv:2502.09443}
}

@inproceedings{wu2025eci,
  title     = {Error-Quantified Conformal Inference for Time Series},
  author    = {Wu, Junxi and Lin, Yilin and Vovk, Vladimir and Chen, Changyou},
  booktitle = {Proceedings of the International Conference on Learning Representations (ICLR)},
  year      = {2025},
  note      = {openreview.net/forum?id=RD9q5vEe1Q}
}

@article{dai2024deepseekmoe,
  title={Deepseekmoe: Towards ultimate expert specialization in mixture-of-experts language models},
  author={Dai, Damai and Deng, Chengqi and Zhao, Chenggang and Xu, RX and Gao, Huazuo and Chen, Deli and Li, Jiashi and Zeng, Wangding and Yu, Xingkai and Wu, Yu and others},
  journal={arXiv preprint arXiv:2401.06066},
  year={2024}
}

@article{jiang2024mixtral,
  title={Mixtral of experts},
  author={Jiang, Albert Q and Sablayrolles, Alexandre and Roux, Antoine and Mensch, Arthur and Savary, Blanche and Bamford, Chris and Chaplot, Devendra Singh and Casas, Diego de las and Hanna, Emma Bou and Bressand, Florian and others},
  journal={arXiv preprint arXiv:2401.04088},
  year={2024}
}

@article{lepikhin2020gshard,
  title={Gshard: Scaling giant models with conditional computation and automatic sharding},
  author={Lepikhin, Dmitry and Lee, HyoukJoong and Xu, Yuanzhong and Chen, Dehao and Firat, Orhan and Huang, Yanping and Krikun, Maxim and Shazeer, Noam and Chen, Zhifeng},
  journal={arXiv preprint arXiv:2006.16668},
  year={2020}
}

@inproceedings{gibbs2021adaptive,
  title={Adaptive conformal inference under distribution shift},
  author={Gibbs, Isaac and Candes, Emmanuel},
  booktitle={NeurIPS},
  year={2021}
}
\bibliographystyle{icml2026}

\newpage
\appendix
\onecolumn
%\section{You \emph{can} have an appendix here.}

\section{Appendix}
We provide extended analysis, implementation details, and algorithmic descriptions to complement the main text. Specifically: 
\begin{itemize} 
\item Section~\ref{sec:appendix_calibration_impl} elaborates on the experimental setup for our calibration evaluations.
\item Section~\ref{sec:appendix_var_analysis} presents extended results including correlation heatmaps and per-expert variance analysis.
\item Section~\ref{subsec:app_ablations} contains additional ablation experiments (resolution of uncertainty estimation and loss function formulation).
\item Section~\ref{sec:appendix_algoirthm} provides the pseudo-code for the MoGU framework, which supplements the PyTorch implementation we provide. 
%\item Section~\ref{sec:appendix_limitations_future} discusses the limitations of our current work and future research directions.
\end{itemize}
\subsection{Experimental Protocols for Calibration Assessment}\label{sec:appendix_calibration_impl}
To evaluate the uncertainty quantification performance of MoGU and the baselines, we employ an Online Conformal Prediction framework that guarantees a 90\% marginal coverage level ($\alpha=0.1$) while adapting to distribution shifts via a sliding window of the most recent 1000 residuals. By restricting calibration to this local history, the method dynamically adapts to non-stationary environments, ensuring that uncertainty estimates reflect the current volatility regime rather than outdated long-term averages.

A critical component of our protocol is the strict prevention of data leakage through a delayed feedback mechanism. Since the ground truth for a horizon $H$ forecast made at time $t$ is only observed at $t+H$, the calibration set is updated strictly with "matured" scores from past predictions for which ground truth has recently become available. Furthermore, to account for variable-specific dynamics and the specific characteristics of each prediction step, we maintain distinct calibration sets for each multivariate channel and for each specific time step within the prediction horizon.

\subsection*{Calibration Methods}

\noindent \textbf{Standard Conformal Prediction:}
Employed for deterministic baselines (trained with MSE). The non-conformity score is defined as the absolute prediction error: $s_t = |y_t - \hat{y}_t|$. We construct symmetric prediction intervals around the point forecast:
\begin{equation*}
    \mathcal{C}_t = [\hat{y}_t - q, \quad \hat{y}_t + q]
\end{equation*}
where $q$ is the $(1-\alpha)$-quantile of the absolute errors computed over the specific calibration window.

\bigskip

\noindent \textbf{Conformalized Quantile Regression (CQR):}
Used for baselines trained to output raw quantiles (via Pinball Loss), predicting base lower ($\hat{y}^{lo}_t$) and upper ($\hat{y}^{hi}_t$) bounds corresponding to the target quantiles $\alpha/2$ and $1 - \alpha/2$, respectively. The non-conformity score measures the maximum violation of these bounds: $s_t = \max(\hat{y}^{lo}_t - y_t, \: y_t - \hat{y}^{hi}_t)$. The calibrated interval corrects the base bounds using the historical violations:
\begin{equation*}
    \mathcal{C}_t = [\hat{y}^{lo}_t - q, \quad \hat{y}^{hi}_t + q]
\end{equation*}
where $q$ denotes the $(1-\alpha)$-quantile of the scores in the calibration set.

\bigskip

\noindent \textbf{Adaptive Conformal Prediction with Variance Scaling (CPVS):}
Employed for MoGU and probabilistic models estimating both mean and standard deviation ($\hat{\sigma}_t$). We use standardized residuals as the non-conformity score: $s_t = |y_t - \hat{y}_t| / \hat{\sigma}_t$. The prediction interval scales the model's predicted uncertainty by the calibrated quantile:
\begin{equation*}
    \mathcal{C}_t = [\hat{y}_t - \hat{\sigma}_t \cdot q, \quad \hat{y}_t + \hat{\sigma}_t \cdot q]
\end{equation*}
Here, $q$ is the $(1-\alpha)$-quantile of the standardized scores in the calibration set.

\subsection{Variance Analysis: Additional Results}\label{sec:appendix_var_analysis}
\subsubsection{Correlation Heatmaps: Uncertainty versus Prediction Error}\label{subsec:appendix_corr_heatmaps}
\begin{figure}[h!]
        \centering
        \begin{subfigure}[t]{0.24\textwidth}
            \centering
            \includegraphics[width=\linewidth]{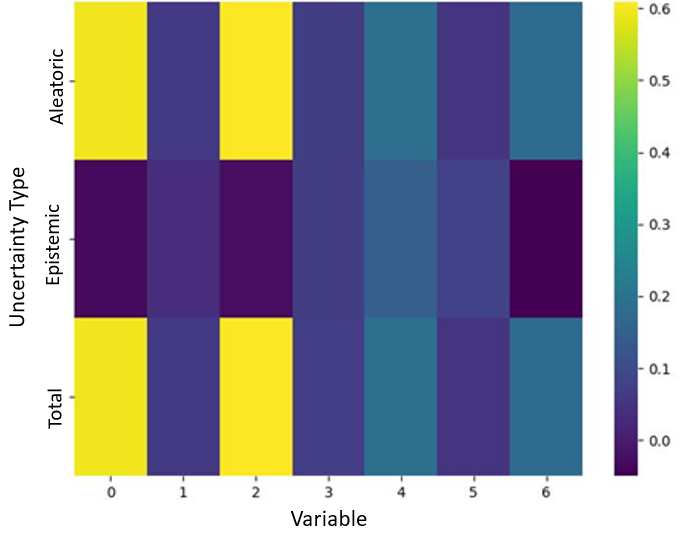}
             \subcaption{ETTh1}
            \label{fig:heatmap_a}
        \end{subfigure}
        \hfill % Adds horizontal space between subfigures
        \begin{subfigure}[t]{0.23\textwidth}
            \centering
            \includegraphics[width=\linewidth]{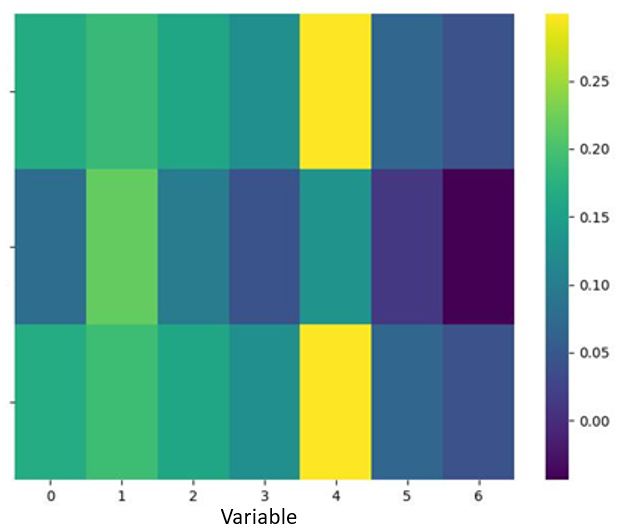}
             \subcaption{ETTh2}
            \label{fig:heatmap_b}
        \end{subfigure}
        \hfill % Adds horizontal space between subfigures
        \begin{subfigure}[t]{0.23\textwidth}
            \centering
            \includegraphics[width=\linewidth]{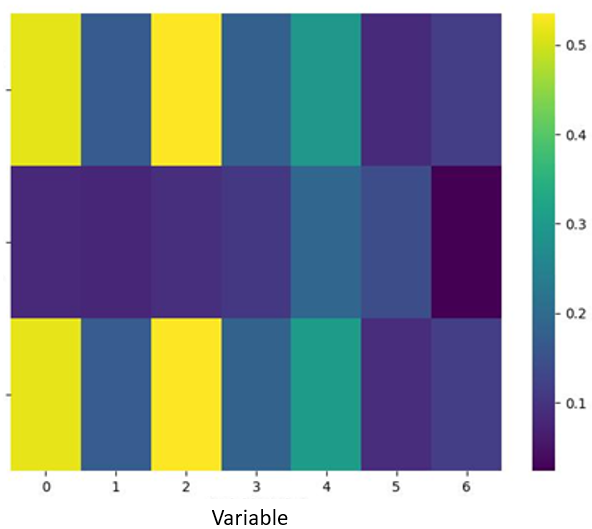}
             \subcaption{ETTm1}
            \label{fig:heatmap_c}
        \end{subfigure}
        \hfill % Adds horizontal space between subfigures
        \begin{subfigure}[t]{0.23\textwidth}
            \centering
            \includegraphics[width=\linewidth]{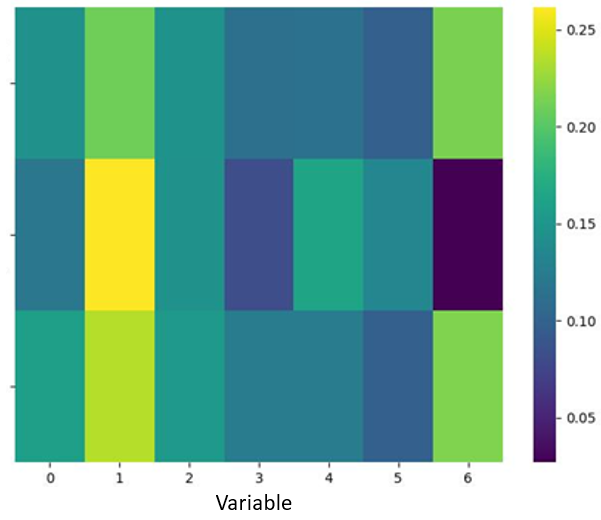}
            \subcaption{ETTm2}
            \label{fig:heatmap_d}
        \end{subfigure}
        
        \caption{Heatmaps of the Pearson correlation between MoGU's reported uncertainties (aleatoric, epistemic, and total) and the MAE of its predictions. The correlation is displayed per variable for the ETT datasets.}
        \label{fig:heatmaps}
    \end{figure}

 The heatmaps in Fig.~\ref{fig:heatmaps} visualize the relationship between the Mean Absolute Error (MAE) of MoGU's predictions and its reported uncertainties (aleatoric, epistemic, and total), when using MoGU with three iTransformer experts. The analysis is presented per variable for each of the ETT datasets, highlighting the extent to which different uncertainty components correlate with predictive error. While the correlation between uncertainty and MAE varies among variables, it remains consistently positive.
\subsubsection{Per-Expert Variance Analysis and Expert Utilization}
\begin{figure}[h!]
    \centering
    \includegraphics [scale=0.60,  trim= 00 300 00 0 ,clip]{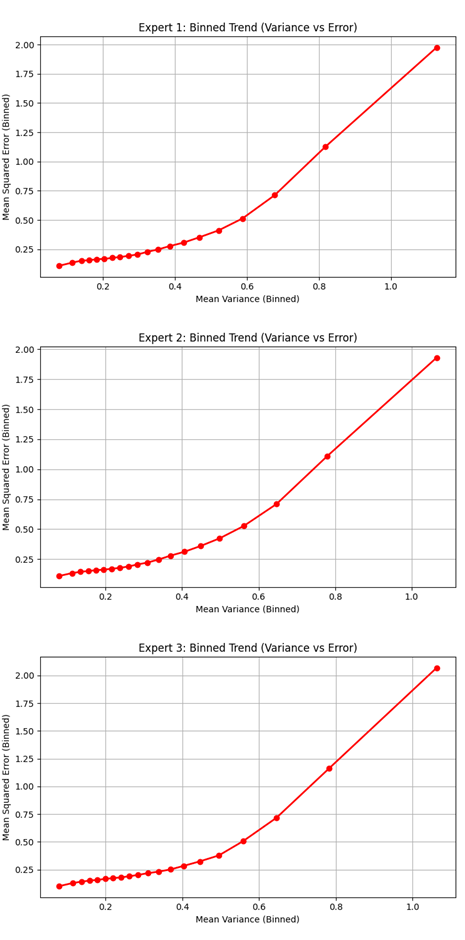}
     \includegraphics [scale=0.60,  trim= 00 150 00 150 ,clip]{image.png}
    \includegraphics [scale=0.60,  trim= 00 00 00 300 ,clip]{image.png}
       \caption{Variance-Error trend analysis for the three experts within the MoGU framework (ETTh1). Data points are grouped into 20 quantile-based bins according to predicted variance. The clear positive correlation between predicted uncertainty and actual squared error demonstrates the reliability of the per-expert uncertainty quantification. (Backbone: iTransformer, Horizon: 96).}
    \label{fig:variance_error_correlation}
\end{figure}

We further investigate the internal dynamics of the MoGU framework by analyzing the distribution of expert weights and the relationship between uncertainty and predictive error. A common failure mode in Mixture-of-Experts (MoE) architectures is expert collapse, where the gating network converges to a "winner-take-all" strategy, over-relying on a single expert. As detailed in Table \ref{tab:expert_weights_distribution}, MoGU maintains a balanced utilization across all experts. %On the ETT datasets, the weights are distributed nearly uniformly (e.g., $\approx 33\%$ each for ETTh1 and ETTm1). %suggesting that the uncertainty-based gating mechanism effectively avoids collapse by dynamically routing samples based on localized expertise. 
Furthermore, we evaluate the fidelity of the internal uncertainty signals by analyzing the relationship between predicted variance and empirical squared error. As illustrated in Figure~\ref{fig:variance_error_correlation}, higher per-expert variance consistently correlates with increased prediction error across all experts. This trend aligns with the global performance of the model and reinforces our calibration analysis, demonstrating that the per-expert uncertainty estimations are not only stable but also serve as reliable indicators of predictive difficulty.
\begin{table}[h]
\centering
\scriptsize % Matches previous results tables
\caption{Analysis of expert weight distribution on ETT datasets. Results report the mean utilization percentage $\pm$ standard deviation across the test set. The near-uniform distribution across all datasets indicates the absence of expert collapse.}
\label{tab:expert_weights_distribution}
\renewcommand{\arraystretch}{1.2}
\setlength{\tabcolsep}{10pt}
\begin{tabular}{lccc}
\toprule
\textbf{Dataset} & \textbf{Expert 1} & \textbf{Expert 2} & \textbf{Expert 3} \\
\midrule
ETTh1 & 33.21\% $\pm$ 0.78\% & 33.80\% $\pm$ 0.51\% & 32.99\% $\pm$ 0.51\% \\
ETTh2 & 36.30\% $\pm$ 2.32\% & 34.29\% $\pm$ 4.03\% & 29.41\% $\pm$ 5.22\% \\
ETTm1 & 33.34\% $\pm$ 2.20\% & 33.35\% $\pm$ 1.58\% & 33.31\% $\pm$ 1.57\% \\
ETTm2 & 30.09\% $\pm$ 5.26\% & 42.62\% $\pm$ 6.99\% & 27.29\% $\pm$ 6.57\% \\
\bottomrule
\end{tabular}
\end{table}

\subsection{Additional Ablations}\label{subsec:app_ablations}

\textbf{Resolution of Uncertainty Estimation.} We provide Table \ref{tab:ablations_time}, discussed in the main text. This table explores an alternative where the expert estimates uncertainty at the variable level ('Time-Fixed'), rather than for each individual time point ('Time-Varying').

\textbf{Loss Function.} We note that the MoGU model can also be optimized through the following MoGE loss:
\begin{equation}\label{eq:mog_loss_alt}
    \mathcal{L}_{}= - \log (\sum_i w_i(x)\mathcal{N}(y;y_i(x), \sigma_i^2(x))) 
\end{equation}
where $\mathcal{N}$ is the Normal density function and the loss
has the form of a Negative Log Likelihood (NLL) of a MoG distribution.
We compare the performance of our model when using the loss presented in Eq. \ref{eq:mog_loss} and when using the aforementioned alternative (Eq. \ref{eq:mog_loss_alt}). The results of this experiment, presented in Table \ref{tab:ablations_loss} in our Appendix, suggest that optimizing with our proposed loss (Eq. \ref{eq:mog_loss}) yields more effective learning and consistently better results by imposing a stricter constraint on expert learning compared to the MoGE loss.

\begin{table*}[h!]
\centering
\scriptsize
\caption{Ablation study on the temporal resolution of reported uncertainty. We compare estimating variance once per horizon (Time-Fixed) versus per time point (Time-Varying). All results use iTransformer with a 96-step horizon. Bold indicates the best result per dataset.}
\label{tab:ablations_time}
\renewcommand{\arraystretch}{1.2}
\setlength{\tabcolsep}{12pt}
\begin{tabular}{l cc cc}
\toprule
\textbf{Dataset} & \multicolumn{2}{c}{\textbf{Time-Fixed Resolution}} & \multicolumn{2}{c}{\textbf{Time-Varying Resolution}} \\
\cmidrule(lr){2-3} \cmidrule(lr){4-5}
\textbf{Metric} & MAE & MSE & MAE & MSE \\
\midrule
ETTh1 & 0.401 & 0.392 & \textbf{0.400} & \textbf{0.380} \\
ETTh2 & 0.337 & 0.290 & \textbf{0.336} & \textbf{0.283} \\
ETTm1 & 0.360 & 0.324 & \textbf{0.356} & \textbf{0.320} \\
ETTm2 & \textbf{0.255} & \textbf{0.174} & 0.260 & 0.179 \\
\bottomrule
\end{tabular}
\end{table*}

\begin{table*}[h!]
\centering
\scriptsize
\caption{Ablation study of the MoGU loss formulation. We compare our proposed loss in Eq.~\ref{eq:mog_loss} against the alternative MoGE formulation in Eq.~\ref{eq:mog_loss_alt} across varying horizons on the ETTh2 dataset.}
\label{tab:ablations_loss}
\renewcommand{\arraystretch}{1.2}
\setlength{\tabcolsep}{12pt}
\begin{tabular}{l cc cc}
\toprule
\textbf{Horizon} & \multicolumn{2}{c}{\textbf{Alt. MoGE Loss (Eq.~\ref{eq:mog_loss_alt})}} & \multicolumn{2}{c}{\textbf{MoGU Loss (Eq.~\ref{eq:mog_loss})}} \\
\cmidrule(lr){2-3} \cmidrule(lr){4-5}
\textbf{Metric} & MAE & MSE & MAE & MSE \\
\midrule
96  & 0.343 & 0.304 & \textbf{0.336} & \textbf{0.283} \\
192 & 0.389 & 0.378 & \textbf{0.387} & \textbf{0.361} \\
336 & \textbf{0.424} & 0.422 & 0.425 & \textbf{0.415} \\
720 & \textbf{0.438} & \textbf{0.421} & 0.442 & \textbf{0.421} \\
\bottomrule
\end{tabular}
\end{table*}
\subsection{MoGU's Algorithm (Pseduo Code)}\label{sec:appendix_algoirthm}
We provide the pseudo code for MoGU in Listing 1 to enhance clarity and supplement our PyTorch implementation.

We implemented MoGU to be highly configurable, so that users can specify the number of experts, the expert architecture, the mixture type (MoE or MoGE) and the gating mechanism.

\begin{algorithm}[!htb]
%\caption{Mixture-of-Gaussians with Uncertainty-based Gating (MoGU)}
\label{alg:mogu}
\begin{algorithmic}[1]

\STATE \textbf{Require:} Training data $X$, labels $y$
\STATE \textbf{Ensure:} Model parameters $\theta$

\FOR{each training epoch}
  \FOR{each mini-batch $\mathcal{B}$}
    \FOR{each sample $x \in \mathcal{B}$}
      \FOR{each expert $i = 1,\dots,k$}
        \STATE Compute expert likelihood 
        $f_i(x) = \mathcal{N}\!\left(y;\, y_i(x,\theta),\, \sigma_i^2(x,\theta)\right).$
        \STATE Compute uncertainty-based gate  
        $w_i(x,\theta) = \frac{\sigma_i^{-2}(x,\theta)}{\sum_j \sigma_j^{-2}(x,\theta)}.$
      \ENDFOR
    \ENDFOR

    \STATE Compute loss  
    $
      \mathcal{L} = \sum_{x \in \mathcal{B}} \sum_{i=1}^k 
         w_i(x)\,\mathcal{L}_{\mathrm{NLLG}}\!\left(y;\, y_i(x),\, \sigma_i^2(x)\right)
    $
    \STATE Update model parameters $\theta$
  \ENDFOR
\ENDFOR

\STATE \textbf{At test time:}
\STATE Prediction:  $\hat{y}(x) = \sum_i w_i(x)\,y_i(x)$
\STATE Predictive variance:  
$$
\sum_i w_i(x)\,\sigma_i^2(x) 
\;+\;
\sum_i w_i(x)\,\bigl(\hat{y}(x)-y_i(x)\bigr)^2
$$

\end{algorithmic}
\end{algorithm}

\end{document}